\pdfoutput=1

\documentclass[11pt]{article}

\usepackage[preprint]{acl}

\usepackage{times}
\usepackage{latexsym}

\usepackage[T1]{fontenc}

\usepackage[utf8]{inputenc}

\usepackage{microtype}

\usepackage{inconsolata}
\usepackage{array}
\usepackage{ragged2e}
\usepackage{graphicx}
\usepackage{makecell}
\usepackage{hhline}
\usepackage{hyperref}

\newcommand{\heart}{\ensuremath\heartsuit}

\newcommand{\changeA}[1]{\textcolor{black}{#1}}

\title{A Synthetic Data Approach for Domain Generalization of NLI Models}

\author{Mohammad Javad Hosseini$^{*}$ \quad Andrey Petrov \quad Alex Fabrikant \quad Annie Louis$^{*}$ \\
        Google Deepmind \\
        \texttt{\{javadh,apetrov,fabrikant,annielouis\}@google.com}
}

\begin{document}
\maketitle
\begingroup\def\thefootnote{*}\footnotetext{Equal contribution.}\endgroup
\begin{abstract}
Natural Language Inference (NLI) remains an important benchmark task for LLMs. NLI datasets are a springboard for transfer learning to other semantic tasks, and NLI models are standard tools for identifying the faithfulness of model-generated text. There are several large scale NLI datasets
today, and models have improved greatly by hill-climbing on these collections. Yet their realistic performance on out-of-distribution/domain data is less well-understood.
We explore the opportunity for synthetic high-quality datasets to adapt NLI models for zero-shot use in downstream applications across new and unseen text domains. We demonstrate a new approach for generating NLI data in diverse domains and lengths, so far not covered by existing training sets. The resulting examples have meaningful premises, 
the hypotheses are formed in creative ways rather than simple edits to a few premise tokens, and the labels have high accuracy. 
We show that models trained on this data ($685$K synthetic examples) have the best generalization to completely new downstream test settings. On the TRUE benchmark, a T5-small model trained with our data improves around $7\%$ on average compared to training on the best alternative dataset. The improvements are more pronounced for smaller models, while still meaningful on a T5 XXL model. We also demonstrate gains on test sets when in-domain training data is augmented with our domain-general synthetic data.

\end{abstract}


\section{Introduction}

Over the past decade, NLI tasks have been critical for benchmarking the representation strengths of our language models. Today, the accuracy on the oft-reported Multi-NLI \cite{mnli} (MNLI) dataset has reached above 92\%\footnote{\url{https://gluebenchmark.com/leaderboard}} and supersedes human-level performance \cite{nangia2019human}. Simultaneously, the practical applications of NLI models has gathered immense attention in fact-checking and source attribution of LLM outputs \cite{honovich-etal-2022-true-evaluating,ais}. For such downstream tasks, model generalization to new domains and data-distributions is critical. We present a method of synthetic data generation to create a \emph{general} dataset with varied but balanced distribution of premise lengths, domains of text, and NLI labels (Table \ref{tab:gnli_examples}), and demonstrate improved accuracy with the new data.

\begin{table}[t]
\centering
\small
\begin{tabular}{| p{0.93\linewidth} |} \hline

\makecell{\bf Domain = essay} \\ \hline
\textcolor{teal}{P: This book does a great job of putting all the different approaches under one roof, so that you can see what other researchers are doing and how they do it.}  \\ 
\textcolor{red}{H: The book covers different research approaches in a single place so you can compare them.}  \\ 
\textcolor{blue}{L: entailment }  \\ \hline

\makecell{\bf Domain = reddit title} \\ \hline
\textcolor{teal}{P: TIL the difference between "literally" and "figuratively". It was so easy to learn, I literally did a backflip.}  \\ 
\textcolor{red}{H: I did not bother learning the difference between "literally" and "figuratively".}  \\ 
\textcolor{blue}{L: contradiction}  \\ \hline

\makecell{\bf Domain = story for kids} \\ \hline
\textcolor{teal}{P: Once upon a time, there was a very special young lady named Cinderella. Her stepmother and stepsisters were very mean to her. But she continued to be kind and helpful.}  \\ 
\textcolor{red}{H: Cinderella was very kind to everyone.}  \\ 
\textcolor{blue}{L: neutral}  \\ \hline

\end{tabular}
\normalsize
\caption{NLI examples in our synthetic data.}
\label{tab:gnli_examples}
\end{table}

MNLI was a first effort to create a multi-domain NLI \emph{training} dataset with examples from 5 genres spanning fiction, formal texts and conversations, and with single-sentence premise/hypothesis texts. Today, the use of models trained from such datasets has expanded well past routine benchmarking into a variety of practical tasks. In downstream problems involving web-scale LLMs, such as fact-checking of social media text, the domains and texts are clearly more diverse. Yet, as a field, we have not fully explored the distribution-general abilities of our models for downstream semantic tasks beyond NLI itself. There are many other NLI training sets: ANLI \cite{nie2019adversarial} for harder reasoning going beyond stylistic elements, and WANLI \cite{liu2022wanli} that generates synthetic examples through worker and AI collaboration and which replicate complex reasoning patterns.

To complement these efforts, we explore 
the opportunity for \textit{synthetic high-quality datasets to adapt NLI models for zero-shot use in downstream applications across new and unseen text domains}.

Our generation covers nearly 40 realistic and distinct domains, ranging from reviews, social media comments, to legal texts, also with varying lengths of the premise text. Our technique employs a chain of LLM tasks tuned to generate high quality, creative premise-hypothesis pairs together with a 3-way NLI label ({\em entailment}, {\em contradiction}, or {\em neutral} \cite{mnli}). A first step generates domain names, the second produces premises of different lengths in these domains, and the final LLM call produces hypotheses and labels conditioned on each premise. We demonstrate how this approaches creates data with a balanced distribution of domains, labels, and premise lengths. 

We fine-tune NLI models on this synthetic data corpus, and present their accuracy on the TRUE factual consistency benchmark \cite{honovich-etal-2022-true-evaluating}, consisting of 11 tasks unseen by our data and other training sets. We show that our \emph{general data}-trained models obtain state-of-the-art NLI performance and single-handedly outperform models trained on MNLI, ANLI, or WANLI with around $7\%$ improvement over the best alternative for T5 \cite{raffel2020exploring} small models. The gap is lower but still around $2\%$ for T5 XXL model size.

Our main contribution is thus that the \emph{general} synthetic data approach improves the generalization power of NLI models, especially when small models and fast inference is key. We further show that, while in-distribution performance is hard to beat for tasks with in-distribution training data, our  synthetic data  can still improve in-distribution performance when used to augment the training data for models with sufficient capacity.

\begin{figure*}[ht]
    \centering
    \includegraphics[width=14cm]{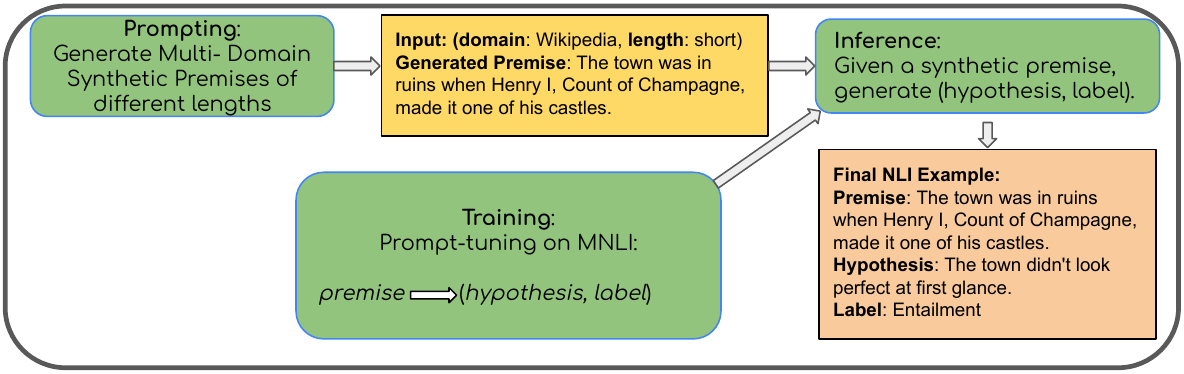}
    \caption{Generating the \emph{General-}NLI examples.}
    \label{fig:concept}
\end{figure*}
\section{Related Work}
\label{sec:related}

We describe various ways of creating NLI data, and why models should generalize beyond the data.

\noindent{\bf Human annotation of NLI examples.} The major datasets
available today were created via costly, time consuming annotation tasks,
and significant human effort.
Standardly, for datasets such as SNLI \cite{bowman-etal-2015-large} and MNLI \cite{mnli}, the annotation starts with a premise sentence
taken verbatim from different sources. Annotators are asked to write a hypothesis sentence that is 
entailed, contradicted, or is neutral to the premise. This writing task is complex for humans, and it is well-known that the collected examples often have undesired stylistic artifacts, for example, the 
hypotheses alone being 
highly predictive of the label \cite{gururangan-etal-2018-annotation}. 
Later efforts (ANLI dataset) have focused on improving the quality of examples by including model adversaries into the annotation rounds \cite{nie2019adversarial}.

The diversity in genre in these datasets depends on the sources from which the premises are drawn. 
The SNLI dataset contains image captions \cite{bowman2015large}. The MNLI dataset \cite{mnli} has 10 domains from the  Open American National
Corpus (OANC)\footnote{\url{https://anc.org/}},
only 5 of which are used for training. The ANLI dataset \cite{nie2019adversarial} contains text from Wikipedia, news, fiction, formal spoken text, and causal or procedural text. In this way, their domain coverage is limited by the chosen corpora and licensing constraints. For example, none of these datasets contain reviews, forum discussions, and science texts, domains which are prevalent and important in applications. 

Understandably, different sources and methods of
data collection produces training examples of a certain style and distribution. 
The generalization of these NLI sets to fully new settings is an interesting problem \cite{adila2022understanding}, yet
less explored. Our work aims to shed some insights here. 

\vspace{2mm}
\noindent{\bf Domain generalization.} This problem of training/test mismatch receives less importance during development
since models are trained and evaluated on splits of the same dataset. But real world applications of these models cannot assume that the test data matches the training distribution. Models need to be adapted to individual test domains using domain adaptation, or alternatively one could train domain-general models which work well on multiple unseen domains or distributions. We focus on this problem of domain generalization. 

There is a large body on work on how training and optimization of models can be 
adapted for better generalization \cite{pmlr-v28-muandet13,domaingeneralsurvey}. Another well-known
approach, especially in computer vision, is to augment the training data to increase its diversity and reduce model overfitting \cite{domainrandomization,multicomponent,novelvisiondomains}. \changeA{\newcite{liu2020empirical} uses data augmentation to overcome multiple adverserial attacks.}
In this work, we explore the usefulness of synthetic data generation for the domain-generalization of NLI models. 

\vspace{2mm}
\noindent{\bf Synthetic NLI data generation.} Today's LLMs have opened up the possibility of synthetic data to aid many NLP tasks \cite{puri-etal-2020-training,gal,agrawal2023qameleon,li-etal-2023-synthetic}.
For NLI, synthetic data has been used for different goals: 
augmenting small training sets and adding \emph{in-domain} 
examples for self-training \cite{vu-etal-2021-strata,gal}, and increasing the size of harder
examples \cite{liu2022wanli}. Most of these methods prompt LLMs with 
sequences of premise-hypothesis sentence pairs.  
The pairs are then labelled by a teacher model. 
\citet{liu2022wanli} specialize the generation towards complex linguistic and reasoning 
patterns in 
existing datasets. We employ synthetic data to improve the diversity 
and balance of training data along the dimensions of domain, premise length,
and label skew.
\section{Synthesizing a \emph{General}-NLI Dataset}
\label{sec:dataset_creation}

We aim to generate NLI examples in different domains, and with premises of varied lengths.\footnote{Longer hypotheses are not of interest typically. A hypothesis is entailed if hypothesis is true given the premise. Long hypotheses are less likely to contain precise entailment and contradiction relations, with some exceptions such as summaries.}

Generating synthetic examples for NLI is in fact a challenging problem. The goal is to produce a pair of texts, which exemplify the reasoning behind different NLI labels. But for the data to be useful, these texts must have creative content and language, and require reasoning. Prior synthetic data generation approaches  \cite{gal,liu2022wanli} generate the
premise and hypothesis sentences as a single sequence,
followed by annotating the label using a teacher model or human raters.  Instead, to achieve maximum control over multiple dimensions: genre, length, and different NLI labels, we 
generate this dataset in two steps: (i) enumerate diverse domains and generate premises in those domains (Section \ref{sec:gen_multi_domain_premises}), (ii) generate hypotheses and labels given the premises (Section \ref{sec:gen_hyp_label}).
Figure \ref{fig:concept} provides an overall view into the LLM tasks we use for generating our data. 

\subsection{Generating Premises with Varied Lengths in Multiple Domains}
\label{sec:gen_multi_domain_premises}

Going beyond distinctions based on the source of a text, it is hard to define the boundaries of a domain in a strict manner. Properties of a text differ along many dimensions: its genre (e.g., news, poetry, or fiction), topic (e.g. politics, science), and the platform or venue for the content, either spoken or written (e.g., reddit, email, image captions, or telephone conversations).
We adopt a practical perspective, and consider all these distinctions as the latent features leading to differences between text collections.
It is in fact well-known that stylistic variations in NLI datasets impact  generalization \cite{BelinkovPSDR19,adila2022understanding}. 

So we do not start with predefined domains. Rather, we first build a text-generation model which generates triples of domain name, text length, and text in the domain. The resulting texts from different domains are collected into our premise set. We build this model using few-shot prompting of FLAN-PaLM2 L (Unicorn)
model \cite{palm2}\footnote{Available from \href{https://ai.google.dev/tutorials/python_quickstart}{https://developers.generativeai.google}.} using texts from a few seed domains. We draw 18 in-prompt examples of varied lengths from 8 domains ({\em news headlines}, {\em news}, {\em shopping reviews}, {\em wikipedia}, {\em movie reviews}, {\em place reviews}, {\em twitter} and {\em reddit post}). The example texts are taken 
from public websites with a few edits if needed.\footnote{This includes news websites (BBC) for {\em news} and {\em news headlines}, e-commerce and review websites (eBay and thetechoutlook.com) for {\em shopping reviews}, Wikipedia, Google Play for {\em movie reviews}, citymaps.uk and top-rated.online websites for {\em place reviews}, X (Twitter) and Reddit.} We provide the full list of seed examples in Appendix \ref{sec:seed_examples}. We also select these texts to be of different lengths. 

\begin{figure*}[ht]
    \centering
    \includegraphics[width=15cm]{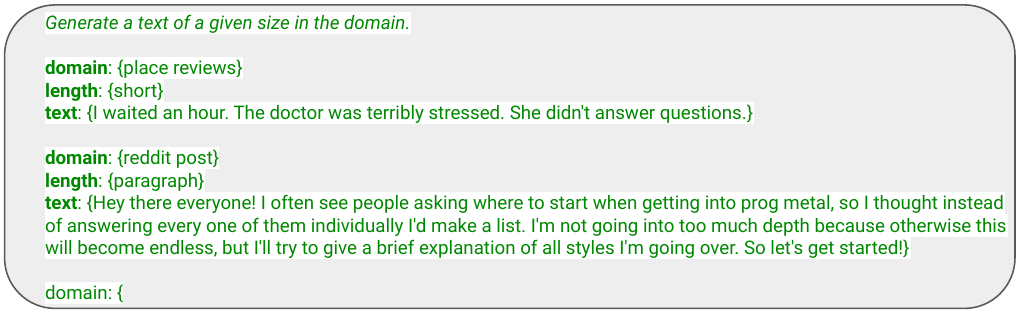}
    \caption{The prompt used to generate new domains. For generating new text, we use the same prompt, add a domain and length category of interest (either {\em short} or {\em paragraph}), and add ``text: \{'' at the end. We take the output up to the first ``\}'' as the generated domain or text.}
    \label{fig:prompt}
    \vspace{-4mm}
\end{figure*}

Figure \ref{fig:prompt} shows our prompt.\footnote{\changeA{We note that we also tried prompting without any instruction (just with few-shot examples) and the generated text had similar quality.}} The length category is set to either {\em short} for single sentences and {\em paragraph} for longer texts. We sample from this model, with a temperature of 1 to get creative \emph{new domains} and texts. Ideally, these samples would directly be useful as premises (with domain and length labels). However, these samples were skewed towards certain domains, e.g., certain types of forums, which neither correspond to real word distributions of web-text, nor serve the purpose of a general model. So we first identify new domains generated by the model. We examined the \emph{new} domains generated in about 1000 samples. Some generated domains were closely related to, or paraphrases of, each other; e.g. both \emph{travel forums} and \emph{US travel forums} were generated. Others were rare or noisy. So we manually selected 38 diverse domains including the seed domain names (Table \ref{tab:domains}).

We then generate balanced samples in these domains, and for the length labels (short, paragraph). We use the same prompt as in Figure \ref{fig:prompt}, but substitute a new domain and length category of interest to it, to generate a text with those properties. This simple text generation model produced high quality and creative texts in different domains. We use these texts as the \emph{premises} in our data.

\begin{table}[t]
\centering
\begin{tabular}{|p{7cm}|}
\hline
ads, blog post, book reviews, casual dialog, chat message, email, essay, fans forum, forum post, google play reviews, government documents, legal, legal document, medical, movie plot, movie reviews, news, news comments, news headlines, phone conversation, place reviews, quora, recipe, reddit comment, reddit title, research paper abstract, scientific article, shopping reviews, song lyrics, sports news, story for kids, student forum, student papers, support forum, travel guides, twitter, wikipedia, youtube comments \\ \hline
\end{tabular}
\caption{Our final list of domains for data generation.}
\label{tab:domains}
\end{table}

\subsection{Generating Hypotheses and Labels}
\label{sec:gen_hyp_label}

We now discuss how we attach hypotheses and labels to our premises to generate complete NLI examples, i.e., (premise, hypothesis, label) triples. 

We train LLMs to leverage existing NLI datasets, and learn the task of writing hypotheses for given premises. 

\begin{figure*}[ht]
    \centering
    \includegraphics[width=15cm]{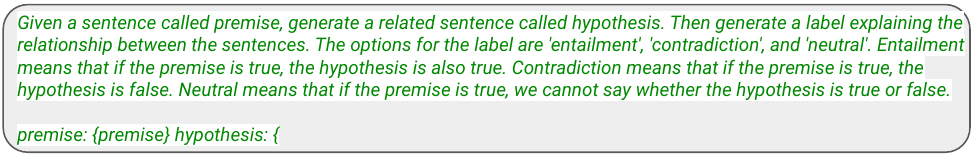}
    \caption{The instruction used for training and inference of the (hypothesis, label) generator model.}
    \label{fig:instruction_pt}
    \vspace{-4mm}
\end{figure*}

Our model conditions on a premise to generate a (hypothesis, label) pair. We generate the label automatically (and accurately, details in next section), and do not need an additional human/teacher labelling step.
This model is trained via prompt-tuning of FLAN-PaLM 540B \cite{chowdhery2023palm,chung2022scaling} 
on the training split of the MNLI dataset. Figure \ref{fig:instruction_pt} shows our prompt which has definitions for the three NLI labels similar to the MNLI annotation guidelines \cite{mnli}.\footnote{\changeA{We also tried another similar instruction for defining the task and labels (different wordings). However, we did not observe meaningful differences with the final prompt shown in the paper. This is probably because the LLM learns the task well after being prompt-tuned on a large set of examples (MNLI).}
}

We used prompt-tuning \cite{lester-etal-2021-power} for training, in lieu of fine-tuning, for two reasons: a) With prompt-tuning, only a few embeddings are updated (100 in our experiments) leading to efficient training, and b) prompt-tuning provides regularization and avoids memorization of the training set details.\footnote{See details of our prompt-tuning running time and hyper-parameters in Appendix \ref{sec:hyper-parameters-palm}.} \changeA{We note that in our preliminary experiments, we also tried just prompting LLMs (no training) to generate NLI examples; however, we did not obtain high quality examples.}

Using the prompt-tuned model, we perform inference once on each of the premises obtained from Section \ref{sec:gen_multi_domain_premises}.\footnote{We discard examples if a) the generated text for hypothesis-premise pair is mis-formatted, or b) the generated labels are not among {\em entailment}, {\em neutral}, and {\em contradiction}. Such errors account for less than $1$\% of the generated data.}
We note that large models and regularization were important for creative generation of hypotheses. A  T5 XL model (3B parameters) fine-tuned on the same task lead to examples with poor creativity and low utility for training. In many cases, the synthetic hypothesis was  a subset of the premise (entailment) or had some simple modifications (e.g., negation) to introduce contradiction.

\subsection{The Final \emph{General} Dataset}

We generated (premise, hypothesis, label) triples in the $38$ domains (from Section \ref{sec:gen_multi_domain_premises}), and for two length categories ({\em short} and {\em paragraph}). The final dataset contains $684,929$ examples. We hold-out $500$ examples for creating a human annotated test set, and split the rest into training, development, and test splits.

Table \ref{tab:splits} shows the data size and label distributions in each split. The number of examples from each label is relatively balanced ($35.4$\% entailment, $31.1$\% contradiction, and $33.5$\% neutral). \changeA{The balanced distribution of the GNLI dataset is predictable given how we generated the data. We use the MNLI data to prompt-tune our generator and this training data has a balanced class distribution. Therefore, after our prompt-tuning, the synthetic GNLI data also has a relatively balanced class distribution. This is in contrast to the previous WANLI dataset that used in-context learning to generate synthetic examples \cite{liu2022wanli}. Although they had balanced in-context examples (1/3 for each label), the final dataset had only $15\%$ contradiction. This shows that prompt-tuning is effective in mirroring the training label distribution. We also note that if the original dataset was not balanced, we could perform sampling to obtain a balanced dataset. Alternatively, we could use a weighted loss function.}

Our generation is also balanced with respect to the premise length and domain by design (we sample examples in a stratified manner). The average number of words per \emph{short} premise is $21$, and $60$ for \emph{paragraph} length. The average number of words in hypotheses is mostly uniform, $10$ and $12$ for \emph{short} and \emph{paragraph} length premises.

Table \ref{tab:examples} shows a few examples from our data. The premises come from different domains and are diverse in form and topic.  
Hypotheses are relevant to the premise and are creative in contrast to slightly modifying the premise and/or taking a subset of it. These attributes are unlike our observations with smaller and less powerful language models such as T5 XL (Section \ref{sec:gen_hyp_label}). In our experiments, we empirically demonstrate the impact of this data for training. 
We note that the idea here is to generate diverse data in different domains. It is possible that some of these examples are not factual, but the truth of the hypothesis is checked against the premise, not any background knowledge.

\begin{table}
\centering
\small
\begin{tabular}{| p{.20\linewidth} | p{.12\linewidth} | p{.51\linewidth} |} \hline
{\bf \textsc{Split}} & {\bf \textsc{Size}} & {\bf \textsc{\# Labels (E/C/N)}}\\ \hline
All & 684,929 & 242,154 / 212,950 / 229,325  \\ \hline
Train & 670,739 & 237,325 / 208,676 / 224,738 \\ \hline
Dev & 6,845 & 2,453 / 2,146 / 2,246 \\ \hline
Test & 6,845 & 2,376 / 2,128 / 2,341 \\ \hline
Human annotated test & 490 & 181 / 155 / 154\\ \hline
\end{tabular}
\caption{Different splits of our \emph{general}-data. \emph{Human annotated test} are 490 (out of 500) examples where at least $2$ out of $3$ annotators have agreed on the label. 
}
\label{tab:splits}
\vspace{-4mm}
\end{table}

\begin{table*}
\centering
\small
\begin{tabular}{| p{.11\linewidth} | p{.44\linewidth} | p{.20\linewidth} | p{.10\linewidth} |} \hline
{\bf \textsc{Domain and Length}} & {\bf \textsc{Premise}} & {\bf \textsc{Hypothesis}} & {\bf \textsc{Label}}\\ \hline
travel guides / short & {\bf This charming boutique offers 43 rooms and suites in the heart of historic St John's}, and is the perfect base for exploring Antigua's rich history & The boutique is located right in the middle of the historic area. & entailment \\ \hline
support forum / short & {\bf I'll be posting a video with the solution} once my phone finishes resetting. & I've already solved the problem. & neutral \\ \hline
legal document / paragraph & This Agreement will bind and inure to the benefit of both parties hereto and their respective personal representatives, heirs, successors, and permitted assigns. {\bf Any attempt by any party hereto to assign, sell or otherwise transfer all or part of his or her rights or obligations under this Agreement, other than as provided herein, will be null and void}, notwithstanding the existence of any provision of law to the contrary. & This agreement allows for any party to reassign their rights and obligations. & contradiction \\ \hline
phone conversation / short & A. What’s better for us for dinner tonight, Italian or Indian? B. Well, Italian is cheaper, but {\bf Indian is quicker to order}. & Ordering Indian food takes a long time but it is better. & contradiction \\ \hline
essay /short & {\bf The first three days of the trip were fantastic.} I had a blast with my friends. & The first three days of the trip were fantastic; the rest was horrible. & neutral \\ \hline

place reviews / short & {\bf The food was fine} but there was only one couple serving that night and it was very busy. & The food tasted like it had been in the microwave for too long. & contradiction \\ \hline
\end{tabular}
\caption{Synthetic examples from our data. We show examples with different domains, length categories, and labels. The most relevant part of the premise is bolded manually for ease of reading.}
\label{tab:examples}
\vspace{-4mm}
\end{table*}

We also performed a human annotation experiment to 
a) to understand the accuracy of labels on our generated examples, and
b) create a curated high-accuracy multi-domain test set for evaluation.

Each of the $500$ generated examples was annotated with an NLI label by three of the paper's authors. Note that the examples were taken verbatim from the models, and are not revised by the annotators. The average Cohen's $\kappa$ score between annotators is $67.97\%$, indicating substantial agreement.
A majority label ($2$ out of $3$ annotators) was obtained in 490 examples, and we use these as a human-annotated test set. We identified a subset, called \emph{unanimous}, where all annotators agreed on the labels ($344$/$500$).  

\changeA{Annotators disagreed with examples that were ambiguous, e.g., the premise did not provide enough context. These cases are usually borderline and challenging. For the following example, one author annotated it as entailment and two authors annotated it as neutral (the person might still have not solved the problem).}

\changeA{{\it premise}: ``I’ll be posting a video with the solution once my phone finishes resetting.''}

\changeA{{\it hypothesis}: ``I’ve already solved the problem''}

We also measured the accuracy of synthetic labels from our model against the \emph{majority} and \emph{unanimous} labels from human annotators. Model labels have high accuracy with $80.41\%$ against majority and $89.53\%$  on the \emph{unanimous} examples. The $\kappa$ coefficient between model labels and majority  and unanimous subsets is also high ($70.53\%$ and $84.17\%$ respectively).

\section{Experiments}

We explore the strengths of our \emph{general}-dataset (GNLI for brevity), by examining the model predictions on data unseen during training. 

We compare models trained on our data with those trained on other large NLI training sets. We choose three such sources: the MNLI dataset ($392$K training examples), ANLI ($162$K) with examples that are harder for MNLI trained models, and WANLI ($10$2K), a dataset created by machine-human collaboration. 
We note that all of these datasets are collected with a similar methodology, i.e., given a premise (and optionally a label), annotators (or LLMs) write a hypothesis. The final label is then manually assigned to the example (if not given as input). In addition, WANLI and GNLI have used MNLI exemplars
(few-shot or supervised learning) for data generation. So these datasets would have similar properties in theory. 

For all these sources, we trained T5 models \cite{raffel2020exploring} as a standard test bed, and explore models of different sizes: small (60M), large (770M), and XXL (11B). We trained on the respective training splits, and tune hyper-parameters on the corresponding validation sets. For WANLI which does not contain a validation split, we used the MNLI validation data. We also trained models on the combination of GNLI and other datasets. For these combined models, we tuned hyper-parameters based on the classification accuracy on the development set of GNLI.\footnote{See hyper-parameter details in Appendix \ref{sec:hyper-parameters-t5}.}

We train two classifiers for each model size (e.g., T5 XXL) and training data (e.g., ANLI): a 3-way classifier with all the three labels, and a binary classifier. For the binary case, we convert each NLI dataset into a binary dataset with entailment and non-entailment (neutral and contradiction) labels. We use the binary classifiers and 3-way classifiers for factual consistency evaluation and NLI benchmarks, respectively.

\setlength{\tabcolsep}{2.5pt}
\begin{table*}[ht!]
\centering
\small
\begin{tabular}{|l|*{12}{m{0.06\linewidth}|}}
\hline 
 & FRANK & QAGS C & QAGS X & MNBM & Summ Eval & BEGIN & Dial Fact & Q$^2$ & PAWS & FEVER & Vitamin C & Avg \\ \hline 
\multicolumn{13}{|c|}{ {\bf \textsc{ T5 small } } } \\ \hline
MNLI & 49.62 & 37.76 & 58.30 & 70.30 & 45.97 & 80.77 & 76.10 & 68.40 & 51.68 & 89.33 & 70.10 & 63.48 \\  \hline
ANLI & 50.95 & 54.64 & 44.70 & 53.99 & 51.34 & 57.66 & 55.39 & 45.02 & 47.09 & 55.24 & 53.50 & 51.77 \\  \hline
WANLI & 57.99 & 54.14 & 70.21 & 69.90 & 48.98 & 65.79 & 77.62 & 68.97 & 51.51 & 84.35 & 67.85 & 65.21 \\  \hline
M + A + W & 50.20 & 47.35 & 61.76 & 69.69 & 46.75 & 79.08 & 76.01 & 66.00 & \textbf{59.15} & 89.94 & 72.71 & 65.33 \\  \hline
GNLI & \textbf{67.32} & \textbf{60.22} & \textbf{72.39} & \textbf{76.91} & \textbf{56.29} & \textbf{82.21} & \textbf{81.23} & \textbf{72.10} & 57.33 & \textbf{90.54} & \textbf{76.11} & \textbf{72.06} \\  \hline 
\multicolumn{13}{|c|}{ {\bf \textsc{ T5 large } } } \\ \hline
MNLI & 79.15 & 58.13 & 79.56 & 79.27 & 61.59 & 82.13 & 87.65 & 77.32 & 75.82 & 93.97 & 81.60 & 77.84 \\  \hline
ANLI & 81.78 & 74.69 & 81.81 & 75.49 & 71.60 & 78.21 & 85.63 & 78.43 & 84.72 & 94.03 & \textbf{89.63} & 81.46 \\  \hline
WANLI & 80.31 & 74.46 & 70.11 & 67.70 & 72.86 & 80.37 & \textbf{89.15} & \textbf{82.16} & 83.17 & 93.82 & 82.79 & 79.72 \\  \hline
M + A + W & 83.57 & 72.28 & 82.27 & 78.28 & 72.61 & 81.13 & 87.25 & 79.81 & \textbf{85.86} & 94.63 & 86.48 & 82.20 \\  \hline
GNLI & \textbf{90.14} & \textbf{81.33} & \textbf{84.02} & \textbf{79.49} & \textbf{79.75} & \textbf{83.45} & 88.76 & 79.77 & 84.63 & \textbf{94.73} & 85.86 & \textbf{84.72} \\  \hline 
\multicolumn{13}{|c|}{ {\bf \textsc{ T5 xxl } } } \\ \hline
MNLI & 88.18 & 79.03 & 83.07 & 78.31 & 72.35 & 81.76 & 88.32 & 76.84 & 83.11 & 95.13 & 84.58 & 82.79 \\  \hline
ANLI & 87.90 & 82.08 & 84.68 & 76.41 & 75.79 & 79.77 & 81.06 & 74.68 & 86.35 & 93.46 & \textbf{90.59} & 82.98 \\  \hline
WANLI & 88.59 & 72.18 & 82.85 & 73.29 & 74.61 & 82.47 & \textbf{92.40} & \textbf{84.88} & 87.29 & 94.97 & 87.38 & 83.72 \\  \hline
M + A + W & 90.60 & \textbf{87.23} & 86.73 & 79.49 & \textbf{79.44} & 83.56 & 85.56 & 76.45 & \textbf{89.95} & 95.07 & 88.55 & 85.69 \\  \hline
GNLI & \textbf{91.38} & 85.48 & \textbf{87.03} & \textbf{79.97} & 79.28 & \textbf{84.00} & 88.57 & 77.40 & 87.10 & \textbf{95.34} & 87.69 & \textbf{85.75} \\  \hline 
\end{tabular}
\caption{
Evaluation of multiple trained models on the TRUE benchmark. \changeA{The rows M + A + W show the results of training models on the mixture of MNLI, ANLI, and WANLI.}
Results are split into three blocks based on the LM size
(T5 small, T5 large, and T5 XXL). We report average AUC-ROC results on all the datasets (expressed as percentages). The best result for each model size and dataset is bolded.
}
\label{tab:true_results}
\end{table*}

\subsection{Performance on Unseen Factual Consistency Benchmarks}
\label{sec:true}

We first test different models on 
data unseen by all of them. We use the TRUE benchmark, a collection of 11 evaluation datasets that contain human annotations for factual consistency in diverse tasks. The tasks include: A) {\bf abstractive summarization}: FRANK \cite{pagnoni2021understanding}, SummEval \cite{fabbri2021summeval}, MNBM \cite{maynez2020faithfulness}, \cite{wang-etal-2020-asking}, QAGS-CNNDM \cite{wang-etal-2020-asking}, and QAGS-XSum \cite{wang-etal-2020-asking}. B) {\bf dialogue generation}: BEGIN \cite{dziri2022evaluating}, Q$^2$ \cite{honovich2021q2}, and DialFact \cite{gupta2022dialfact}. C) {\bf fact verification}: FEVER \cite{thorne-etal-2018-fact}, and VitaminC \cite{schuster-etal-2021-get}. D) {\bf paraphrase detection}: PAWS \cite{zhang-etal-2019-paws}. The benchmark standardizes the above datasets by converting all annotations to binary labels corresponding to whether the entire text is factuality consistent w.r.t the grounding text or not.
This task is a downstream application of NLI models and importantly, the data in this benchmark was not created using the same protocol as NLI benchmarks.

\changeA{We train different sizes of T5 models on MNLI, ANLI, WANLI, and GNLI. In addition, we report results on models trained on the mixture of MNLI, ANLI, and WANLI (M + A + W). Table \ref{tab:true_results} shows the results.} Following previous work, we report the Receiver Operating Characteristic Area Under the Curve (ROC AUC) for binary detection of inconsistent examples. GNLI outperforms MNLI on all datasets across model sizes showing that it has much stronger generalization. On average, GNLI outperforms MNLI for T5-small trained models by a $8.58\%$ margin, T5-large trained models by $6.88\%$, and T5-XXL trained models by $2.96\%$. 

To the best of our knowledge, the previously reported best NLI model on the TRUE benchmark was T5 XXL trained on ANLI \cite{honovich-etal-2022-true-evaluating,gekhman2023trueteacher}. GNLI obtains $6.85\%$ improvement on average over the best alternative with a single dataset for T5 small (WANLI), $3.26\%$ for T5 large (ANLI), and $2.03\%$ for T5 XXL (WANLI). Therefore, GNLI obtains a new state-of-the-art result on TRUE, outperforming other models with large margins on average and on almost all of the individual test sets within TRUE. \changeA{In addition, GNLI alone outperforms the mixture of MNLI, ANLI, and WANLI by a relatively large margin for T5 small and T5 large, and by a small margin for T5 XXL.} 

We note that \newcite{gekhman2023trueteacher} have recently proposed a synthetic dataset, called TrueTeacher, for the task of factual consistency detection. They used summarization models to condense CNN/DM articles, and labelled the document-summary pair with FLAN-PaLM 540B \cite{chowdhery2023palm} based NLI model. Models trained on the TrueTeacher outperformed ANLI ones. We trained a T5 XXL on the TrueTeacher data and observed better performance compared to GNLI as well ($88.06\%$ vs $85.75\%$ on average). This is expected since TrueTeacher is collected directly for the task of factual consistency detection and makes only the binary distinction.

\begin{figure*}[ht!]
    \centering
    \includegraphics[width=16cm]{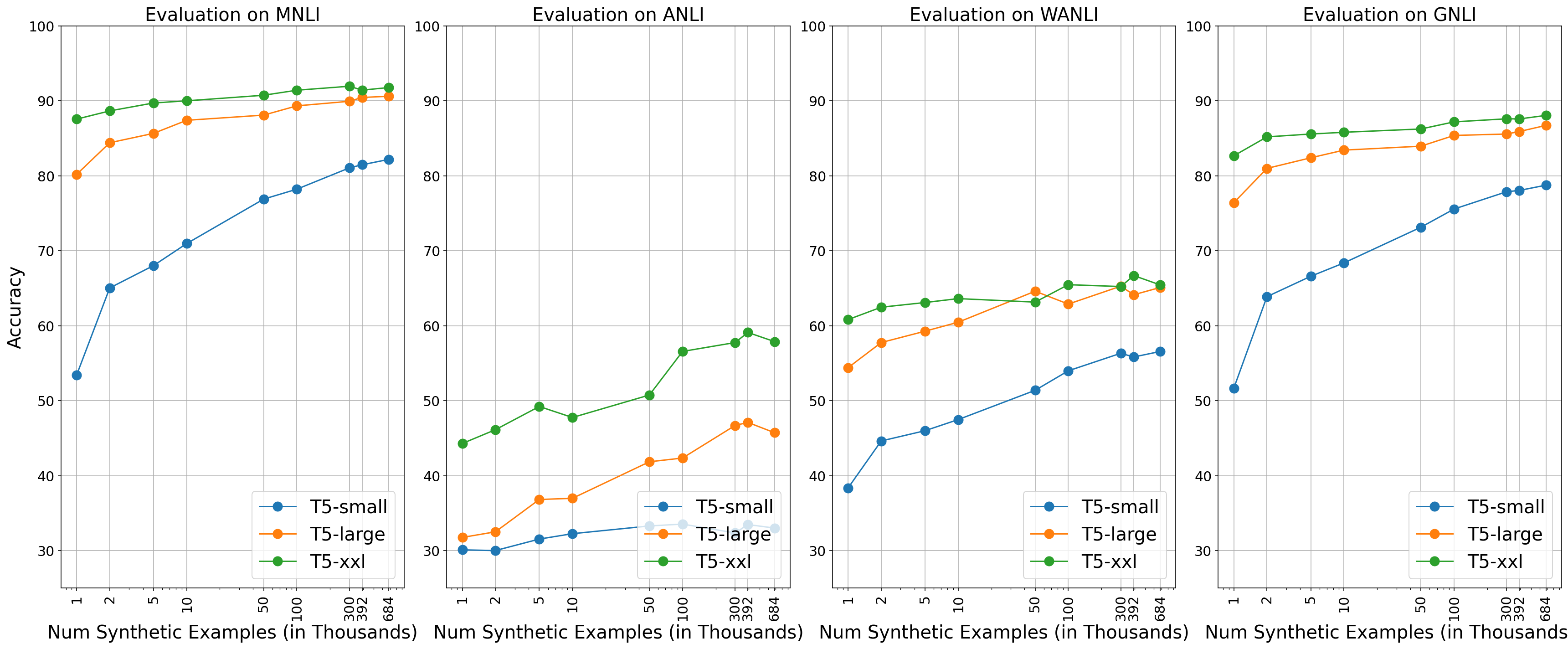}
    \caption{Accuracy of different T5 models when trained on different number of training examples from GNLI. Each plot has the results on one evaluation set.}
    \label{fig:synthetic_plot}
    \vspace{-4mm}
\end{figure*}

\begin{table}[ht!]
\centering
\small
\begin{tabular}{|m{0.27\linewidth}|*{4}{m{0.12\linewidth}|}}
\hline 
 Train / Eval & MNLI & ANLI & WANLI & GNLI Human \\ \hline 
\multicolumn{5}{|c|}{ {\bf \textsc{ T5 small } } } \\ \hline
MNLI & \textbf{83.37} & 31.34 & 56.52 & 75.31 \\
ANLI & 70.35 & \textbf{48.31} & 52.70 & 67.14 \\
WANLI & 60.40 & 36.41 & \textbf{72.60} & 57.76 \\
GNLI  & 82.18 & 33.00 & 56.56 & \textbf{77.14} \\ \hline
MNLI + GNLI & 82.66 & 30.94 & 55.82 & \underline{77.76} \\
ANLI + GNLI & \underline{72.89} & 37.94 & 48.65 & \underline{69.80} \\
WANLI + GNLI & \underline{78.02} & 34.87 & \underline{86.69} & \underline{76.53} \\ \hline
\multicolumn{5}{|c|}{ {\bf \textsc{ T5 large } } } \\ \hline
MNLI & \textbf{90.83} & 40.22 & 63.58 & 82.24 \\
ANLI & 86.87 & \textbf{63.62} & 63.70 & 81.22 \\
WANLI & 81.10 & 48.03 & \textbf{92.08} & 77.14 \\
GNLI  & 90.61 & 45.72 & 65.10 & \textbf{83.67} \\ \hline
MNLI + GNLI & \underline{91.04} & \underline{44.94} & \underline{65.54} & \underline{82.45} \\
ANLI + GNLI & \underline{90.61} & \underline{63.69} & \underline{65.03} & \underline{83.67} \\
WANLI + GNLI & \underline{87.80} & \underline{48.34} & \underline{96.84} & \underline{83.88} \\ \hline
\multicolumn{5}{|c|}{ {\bf \textsc{ T5 xxl } } } \\ \hline
MNLI & \textbf{92.11} & 55.44 & 65.92 & \textbf{83.27} \\
ANLI & 90.01 & \textbf{73.37} & 67.04 & 82.86 \\
WANLI & 84.61 & 60.00 & \textbf{86.46} & 81.84 \\
GNLI  & 91.77 & 57.87 & 65.43 & 82.65 \\ \hline
MNLI + GNLI & \underline{92.13} & \underline{55.94} & \underline{67.06} & \underline{84.08} \\
ANLI + GNLI & \underline{91.96} & 72.94 & 66.26 & \underline{84.08} \\
WANLI + GNLI & \underline{90.34} & \underline{60.19} & \underline{87.90} & \underline{84.69} \\ \hline
\end{tabular}
\caption{
Performance on NLI benchmarks (accuracy percentage). The models were trained on the respective datasets and tested on all their own and other datasets' test (or validation) sets. Results are split into three blocks based on the LM size (T5 small,
large, T5 XXL).  We also report the results of combining GNLI with MNLI,
ANLI, and WANLI. We bold the highest accuracy per model size and evaluation dataset, for models trained on the single datasets. For each combined training set (GNLI + X) and model size, if the result is better than the original dataset (X), the number is underlined.
}
\label{tab:nli-3way}
\vspace{-4mm}
\end{table}

\subsection{Cross-Dataset Performance on NLI Benchmarks}
\label{sec:3-way}

We now examine the models on test sets available with large NLI collections (or validation sets in the absence of test data with labels (MNLI and ANLI)). In this case, at least one model has been trained on data from the same test distribution. Here we seek to understand how general our current datasets are and we also include GNLI in this analysis.

Table \ref{tab:nli-3way} shows the results. While all these datasets have been created with the aim of being domain-general, we see that generally the training data distribution makes a huge difference. The best test numbers are usually obtained by training on the corresponding training sets. For example, the best MNLI numbers are obtained with a model that is trained on MNLI. Note that the GNLI dataset (while including a component that is trained on MNLI) does not include the MNLI examples. These results indicate that the style and properties of different NLI test sets are still rather specific to the individual NLI dataset, and a large dataset with the same type of examples performs best on the corresponding test set. 
However, the GNLI dataset is as accurate as MNLI on these test sets, even without the explicit addition of MNLI examples.

We also trained models on all datasets combined with GNLI. We note that in most cases (except for T5 small), the combined datasets have at least some modest improvements over the original datasets (underline numbers in the table).
We speculate that T5 small's model capacity is not high enough to capture all the information in the combined datasets, but once the model capacity increases, we generally see improvements by adding GNLI to the other NLI datasets.

\subsection{How Much Data is Needed for Successful Training?}
\label{sec:data_size}

GNLI is generated synthetically which is a more efficient and cheaper process compared to crowd-annotated data. It is possible to generate as many examples as necessary, and it is unknown in advance how many examples are needed to get a good performance. On the other hand, generating large sets of examples uses more computing resources. In this section, we study the effect of training data size on evaluation accuracy. We sample $N$ thousand synthetic training examples from GNLI, where $ N \in \{1, 2, 5, 10, 50, 100, 300, 392, 671\}$ ($671$K is the full GNLI dataset). We then train T5 models on all these sample sizes.
We then evaluate the trained models on different NLI datasets. The evaluation is on  validation sets from MNLI and ANLI, and  WANLI and GNLI (synthetic) test sets.

Figure \ref{fig:synthetic_plot} shows the results. We observe that in most cases (model sizes and evaluation sets), at least around  $300$K examples is needed to get a decent performance. We also explicitly tested GNLI with $392$K which is the same size as MNLI. In all cases, GNLI $392$K has a very similar accuracy to the full dataset. We also observed similar trends for the TRUE benchmark.

\section{Conclusion}

A decade of increasingly useful NLI benchmarks and datasets have been instrumental in improving LLMs for various tasks. We have presented a new exploration of how the data distribution of each data source still impacts downstream performance on new examples. We proposed a synthetic data approach to mitigate these effects with examples balanced for domain, length and labels. We show that, by drawing on an LLM's parametric knowledge of a broad range of domains, such synthetic data enables us to both train significantly more domain-general NLI models, and to improve intrinsic NLI model performance on in-domain test data by augmenting in-domain training data.

\section{Limitations}
We do not release the synthetic \emph{general} NLI data with our paper. However, our method for generating them can be replicated with access to an LLM, either to directly reproduce our results or to apply our approach to other domains, text lengths, and/or training set sizes of interest. Our process for generation requires multiple LLM tasks which uses more compute than a single stage one. But we note that the data is of high linguistic quality with this method. At the same time, the generated premises could potentially contain fictional information, and should not be used for training models that learn facts from data. We have applied our approach to generalize only one dataset (MNLI), which has examples in English. While we obtain positive results, the results on other datasets, and for other languages remain an empirical question. \changeA{In addition, we performed experiments with FLAN-PaLM 540B (for prompt-tuning) and FLAN-PaLM2 L (for prompting). However, our method is straightforward and can be easily replicated by other LLMs. We expect comparable LLMs should lead to similar results.}

\bibliography{anthology,esda}

\begin{thebibliography}{38}
\expandafter\ifx\csname natexlab\endcsname\relax\def\natexlab#1{#1}\fi

\bibitem[{Adila and Kang(2022)}]{adila2022understanding}
Dyah Adila and Dongyeop Kang. 2022.
\newblock Understanding out-of-distribution: A perspective of data dynamics.
\newblock In \emph{I (Still) Can't Believe It's Not Better! Workshop at NeurIPS
  2021}, pages 1--8. PMLR.

\bibitem[{Agrawal et~al.(2023)Agrawal, Alberti, Huot, Maynez, Ma, Ruder,
  Ganchev, Das, and Lapata}]{agrawal2023qameleon}
Priyanka Agrawal, Chris Alberti, Fantine Huot, Joshua Maynez, Ji~Ma, Sebastian
  Ruder, Kuzman Ganchev, Dipanjan Das, and Mirella Lapata. 2023.
\newblock Qameleon: Multilingual qa with only 5 examples.
\newblock \emph{Transactions of the Association for Computational Linguistics},
  11:1754.

\bibitem[{Belinkov et~al.(2019)Belinkov, Poliak, Shieber, Durme, and
  Rush}]{BelinkovPSDR19}
Yonatan Belinkov, Adam Poliak, Stuart~M. Shieber, Benjamin~Van Durme, and
  Alexander~M. Rush. 2019.
\newblock Don't take the premise for granted: Mitigating artifacts in natural
  language inference.
\newblock In \emph{Proceedings of the 57th Conference of the Association for
  Computational Linguistics, {ACL} 2019, Florence, Italy, July 28- August 2,
  2019, Volume 1: Long Papers}, pages 877--891.

\bibitem[{Bowman et~al.(2015{\natexlab{a}})Bowman, Angeli, Potts, and
  Manning}]{bowman-etal-2015-large}
Samuel~R. Bowman, Gabor Angeli, Christopher Potts, and Christopher~D. Manning.
  2015{\natexlab{a}}.
\newblock \href {https://doi.org/10.18653/v1/D15-1075} {A large annotated
  corpus for learning natural language inference}.
\newblock In \emph{Proceedings of the 2015 Conference on Empirical Methods in
  Natural Language Processing}, pages 632--642, Lisbon, Portugal. Association
  for Computational Linguistics.

\bibitem[{Bowman et~al.(2015{\natexlab{b}})Bowman, Angeli, Potts, and
  Manning}]{bowman2015large}
Samuel~R Bowman, Gabor Angeli, Christopher Potts, and Christopher~D Manning.
  2015{\natexlab{b}}.
\newblock A large annotated corpus for learning natural language inference.
\newblock In \emph{Conference on Empirical Methods in Natural Language
  Processing, EMNLP 2015}, pages 632--642. Association for Computational
  Linguistics (ACL).

\bibitem[{Chowdhery et~al.(2023)Chowdhery, Narang, Devlin, Bosma, Mishra,
  Roberts, Barham, Chung, Sutton, Gehrmann et~al.}]{chowdhery2023palm}
Aakanksha Chowdhery, Sharan Narang, Jacob Devlin, Maarten Bosma, Gaurav Mishra,
  Adam Roberts, Paul Barham, Hyung~Won Chung, Charles Sutton, Sebastian
  Gehrmann, et~al. 2023.
\newblock Palm: Scaling language modeling with pathways.
\newblock \emph{Journal of Machine Learning Research}, 24(240):1--113.

\bibitem[{Chung et~al.(2022)Chung, Hou, Longpre, Zoph, Tay, Fedus, Li, Wang,
  Dehghani, Brahma et~al.}]{chung2022scaling}
Hyung~Won Chung, Le~Hou, Shayne Longpre, Barret Zoph, Yi~Tay, William Fedus,
  Eric Li, Xuezhi Wang, Mostafa Dehghani, Siddhartha Brahma, et~al. 2022.
\newblock Scaling instruction-finetuned language models.
\newblock \emph{arXiv e-prints}, pages arXiv--2210.

\bibitem[{Dziri et~al.(2022)Dziri, Rashkin, Linzen, and
  Reitter}]{dziri2022evaluating}
Nouha Dziri, Hannah Rashkin, Tal Linzen, and David Reitter. 2022.
\newblock Evaluating attribution in dialogue systems: The begin benchmark.
\newblock \emph{Transactions of the Association for Computational Linguistics},
  10:1066--1083.

\bibitem[{Fabbri et~al.(2021)Fabbri, Kry{\'s}ci{\'n}ski, McCann, Xiong, Socher,
  and Radev}]{fabbri2021summeval}
Alexander~R Fabbri, Wojciech Kry{\'s}ci{\'n}ski, Bryan McCann, Caiming Xiong,
  Richard Socher, and Dragomir Radev. 2021.
\newblock Summeval: Re-evaluating summarization evaluation.
\newblock \emph{Transactions of the Association for Computational Linguistics},
  9:391--409.

\bibitem[{Gekhman et~al.(2023)Gekhman, Herzig, Aharoni, Elkind, and
  Szpektor}]{gekhman2023trueteacher}
Zorik Gekhman, Jonathan Herzig, Roee Aharoni, Chen Elkind, and Idan Szpektor.
  2023.
\newblock Trueteacher: Learning factual consistency evaluation with large
  language models.
\newblock \emph{arXiv preprint arXiv:2305.11171}.

\bibitem[{Google~and et~al.(2023)Google~and, Dai, Firat, Johnson, Lepikhin,
  Passos, Shakeri, Taropa, Bailey, Chen, Chu, Clark, Shafey, Huang,
  Meier-Hellstern, Mishra, Moreira, Omernick, Robinson, Ruder, Tay, Xiao, Xu,
  Zhang, Abrego, Ahn, Austin, Barham, Botha, Bradbury, Brahma, Brooks, Catasta,
  Cheng, Cherry, Choquette-Choo, Chowdhery, Crepy, Dave, Dehghani, Dev, Devlin,
  Díaz, Du, Dyer, Feinberg, Feng, Fienber, Freitag, Garcia, Gehrmann,
  Gonzalez, Gur-Ari, Hand, Hashemi, Hou, Howland, Hu, Hui, Hurwitz, Isard,
  Ittycheriah, Jagielski, Jia, Kenealy, Krikun, Kudugunta, Lan, Lee, Lee, Li,
  Li, Li, Li, Li, Lim, Lin, Liu, Liu, Maggioni, Mahendru, Maynez, Misra,
  Moussalem, Nado, Nham, Ni, Nystrom, Parrish, Pellat, Polacek, Polozov, Pope,
  Qiao, Reif, Richter, Riley, Ros, Roy, Saeta, Samuel, Shelby, Slone, Smilkov,
  So, Sohn, Tokumine, Valter, Vasudevan, Vodrahalli, Wang, Wang, Wang, Wang,
  Wieting, Wu, Xu, Xu, Xue, Yin, Yu, Zhang, Zheng, Zheng, Zhou, Zhou, Petrov,
  and Wu}]{palm2}
Rohan~Anil Google~and, Andrew~M. Dai, Orhan Firat, Melvin Johnson, Dmitry
  Lepikhin, Alexandre Passos, Siamak Shakeri, Emanuel Taropa, Paige Bailey,
  Zhifeng Chen, Eric Chu, Jonathan~H. Clark, Laurent~El Shafey, Yanping Huang,
  Kathy Meier-Hellstern, Gaurav Mishra, Erica Moreira, Mark Omernick, Kevin
  Robinson, Sebastian Ruder, Yi~Tay, Kefan Xiao, Yuanzhong Xu, Yujing Zhang,
  Gustavo~Hernandez Abrego, Junwhan Ahn, Jacob Austin, Paul Barham, Jan Botha,
  James Bradbury, Siddhartha Brahma, Kevin Brooks, Michele Catasta, Yong Cheng,
  Colin Cherry, Christopher~A. Choquette-Choo, Aakanksha Chowdhery, Clément
  Crepy, Shachi Dave, Mostafa Dehghani, Sunipa Dev, Jacob Devlin, Mark Díaz,
  Nan Du, Ethan Dyer, Vlad Feinberg, Fangxiaoyu Feng, Vlad Fienber, Markus
  Freitag, Xavier Garcia, Sebastian Gehrmann, Lucas Gonzalez, Guy Gur-Ari,
  Steven Hand, Hadi Hashemi, Le~Hou, Joshua Howland, Andrea Hu, Jeffrey Hui,
  Jeremy Hurwitz, Michael Isard, Abe Ittycheriah, Matthew Jagielski, Wenhao
  Jia, Kathleen Kenealy, Maxim Krikun, Sneha Kudugunta, Chang Lan, Katherine
  Lee, Benjamin Lee, Eric Li, Music Li, Wei Li, YaGuang Li, Jian Li, Hyeontaek
  Lim, Hanzhao Lin, Zhongtao Liu, Frederick Liu, Marcello Maggioni, Aroma
  Mahendru, Joshua Maynez, Vedant Misra, Maysam Moussalem, Zachary Nado, John
  Nham, Eric Ni, Andrew Nystrom, Alicia Parrish, Marie Pellat, Martin Polacek,
  Alex Polozov, Reiner Pope, Siyuan Qiao, Emily Reif, Bryan Richter, Parker
  Riley, Alex~Castro Ros, Aurko Roy, Brennan Saeta, Rajkumar Samuel, Renee
  Shelby, Ambrose Slone, Daniel Smilkov, David~R. So, Daniel Sohn, Simon
  Tokumine, Dasha Valter, Vijay Vasudevan, Kiran Vodrahalli, Xuezhi Wang,
  Pidong Wang, Zirui Wang, Tao Wang, John Wieting, Yuhuai Wu, Kelvin Xu, Yunhan
  Xu, Linting Xue, Pengcheng Yin, Jiahui Yu, Qiao Zhang, Steven Zheng,
  Ce~Zheng, Weikang Zhou, Denny Zhou, Slav Petrov, and Yonghui Wu. 2023.
\newblock \href {http://arxiv.org/abs/2305.10403} {Palm 2 technical report}.

\bibitem[{Gupta et~al.(2022)Gupta, Wu, Liu, and Xiong}]{gupta2022dialfact}
Prakhar Gupta, Chien-Sheng Wu, Wenhao Liu, and Caiming Xiong. 2022.
\newblock Dialfact: A benchmark for fact-checking in dialogue.
\newblock In \emph{Proceedings of the 60th Annual Meeting of the Association
  for Computational Linguistics (Volume 1: Long Papers)}, pages 3785--3801.

\bibitem[{Gururangan et~al.(2018)Gururangan, Swayamdipta, Levy, Schwartz,
  Bowman, and Smith}]{gururangan-etal-2018-annotation}
Suchin Gururangan, Swabha Swayamdipta, Omer Levy, Roy Schwartz, Samuel Bowman,
  and Noah~A. Smith. 2018.
\newblock \href {https://doi.org/10.18653/v1/N18-2017} {Annotation artifacts in
  natural language inference data}.
\newblock In \emph{Proceedings of the 2018 Conference of the North {A}merican
  Chapter of the Association for Computational Linguistics: Human Language
  Technologies, Volume 2 (Short Papers)}, pages 107--112, New Orleans,
  Louisiana. Association for Computational Linguistics.

\bibitem[{He et~al.(2022)He, Nassar, Kiros, Haffari, and Norouzi}]{gal}
Xuanli He, Islam Nassar, Jamie Kiros, Gholamreza Haffari, and Mohammad Norouzi.
  2022.
\newblock {Generate, Annotate, and Learn: NLP with Synthetic Text}.
\newblock \emph{Transactions of the Association for Computational Linguistics},
  10:826--842.

\bibitem[{Honovich et~al.(2022)Honovich, Aharoni, Herzig, Taitelbaum,
  Kukliansy, Cohen, Scialom, Szpektor, Hassidim, and
  Matias}]{honovich-etal-2022-true-evaluating}
Or~Honovich, Roee Aharoni, Jonathan Herzig, Hagai Taitelbaum, Doron Kukliansy,
  Vered Cohen, Thomas Scialom, Idan Szpektor, Avinatan Hassidim, and Yossi
  Matias. 2022.
\newblock \href {https://doi.org/10.18653/v1/2022.naacl-main.287} {{TRUE}:
  Re-evaluating factual consistency evaluation}.
\newblock In \emph{Proceedings of the 2022 Conference of the North American
  Chapter of the Association for Computational Linguistics: Human Language
  Technologies}, pages 3905--3920, Seattle, United States. Association for
  Computational Linguistics.

\bibitem[{Honovich et~al.(2021)Honovich, Choshen, Aharoni, Neeman, Szpektor,
  and Abend}]{honovich2021q2}
Or~Honovich, Leshem Choshen, Roee Aharoni, Ella Neeman, Idan Szpektor, and Omri
  Abend. 2021.
\newblock Q2:: Evaluating factual consistency in knowledge-grounded dialogues
  via question generation and question answering.
\newblock In \emph{Proceedings of the 2021 Conference on Empirical Methods in
  Natural Language Processing}, pages 7856--7870.

\bibitem[{Lester et~al.(2021)Lester, Al-Rfou, and
  Constant}]{lester-etal-2021-power}
Brian Lester, Rami Al-Rfou, and Noah Constant. 2021.
\newblock \href {https://doi.org/10.18653/v1/2021.emnlp-main.243} {The power of
  scale for parameter-efficient prompt tuning}.
\newblock In \emph{Proceedings of the 2021 Conference on Empirical Methods in
  Natural Language Processing}, pages 3045--3059, Online and Punta Cana,
  Dominican Republic. Association for Computational Linguistics.

\bibitem[{Li et~al.(2023)Li, Zhu, Lu, and Yin}]{li-etal-2023-synthetic}
Zhuoyan Li, Hangxiao Zhu, Zhuoran Lu, and Ming Yin. 2023.
\newblock Synthetic data generation with large language models for text
  classification: Potential and limitations.
\newblock In \emph{Proceedings of the 2023 Conference on Empirical Methods in
  Natural Language Processing}, pages 10443--10461.

\bibitem[{Liu et~al.(2022)Liu, Swayamdipta, Smith, and Choi}]{liu2022wanli}
Alisa Liu, Swabha Swayamdipta, Noah~A Smith, and Yejin Choi. 2022.
\newblock Wanli: Worker and ai collaboration for natural language inference
  dataset creation.
\newblock In \emph{Findings of the Association for Computational Linguistics:
  EMNLP 2022}, pages 6826--6847.

\bibitem[{Liu et~al.(2020)Liu, Xin, Ding, Chang, and Sui}]{liu2020empirical}
Tianyu Liu, Zheng Xin, Xiaoan Ding, Baobao Chang, and Zhifang Sui. 2020.
\newblock An empirical study on model-agnostic debiasing strategies for robust
  natural language inference.
\newblock In \emph{Proceedings of the 24th Conference on Computational Natural
  Language Learning}, pages 596--608.

\bibitem[{Maynez et~al.(2020)Maynez, Narayan, Bohnet, and
  McDonald}]{maynez2020faithfulness}
Joshua Maynez, Shashi Narayan, Bernd Bohnet, and Ryan McDonald. 2020.
\newblock On faithfulness and factuality in abstractive summarization.
\newblock In \emph{Proceedings of the 58th Annual Meeting of the Association
  for Computational Linguistics}, pages 1906--1919.

\bibitem[{Muandet et~al.(2013)Muandet, Balduzzi, and
  Schölkopf}]{pmlr-v28-muandet13}
Krikamol Muandet, David Balduzzi, and Bernhard Schölkopf. 2013.
\newblock Domain generalization via invariant feature representation.
\newblock In \emph{Proceedings of the 30th International Conference on Machine
  Learning}, volume~28 of \emph{Proceedings of Machine Learning Research},
  pages 10--18. PMLR.

\bibitem[{Nangia and Bowman(2019)}]{nangia2019human}
Nikita Nangia and Samuel Bowman. 2019.
\newblock Human vs. muppet: A conservative estimate of human performance on the
  glue benchmark.
\newblock In \emph{Proceedings of the 57th Annual Meeting of the Association
  for Computational Linguistics}, pages 4566--4575.

\bibitem[{Nie et~al.(2019)Nie, Williams, Dinan, Bansal, Weston, and
  Kiela}]{nie2019adversarial}
Yixin Nie, Adina Williams, Emily Dinan, Mohit Bansal, Jason Weston, and Douwe
  Kiela. 2019.
\newblock Adversarial nli: A new benchmark for natural language understanding.
\newblock \emph{arXiv preprint arXiv:1910.14599}.

\bibitem[{Pagnoni et~al.(2021)Pagnoni, Balachandran, and
  Tsvetkov}]{pagnoni2021understanding}
Artidoro Pagnoni, Vidhisha Balachandran, and Yulia Tsvetkov. 2021.
\newblock Understanding factuality in abstractive summarization with frank: A
  benchmark for factuality metrics.
\newblock In \emph{Proceedings of the 2021 Conference of the North American
  Chapter of the Association for Computational Linguistics: Human Language
  Technologies}, pages 4812--4829.

\bibitem[{Puri et~al.(2020)Puri, Spring, Shoeybi, Patwary, and
  Catanzaro}]{puri-etal-2020-training}
Raul Puri, Ryan Spring, Mohammad Shoeybi, Mostofa Patwary, and Bryan Catanzaro.
  2020.
\newblock \href {https://doi.org/10.18653/v1/2020.emnlp-main.468} {Training
  question answering models from synthetic data}.
\newblock In \emph{Proceedings of the 2020 Conference on Empirical Methods in
  Natural Language Processing (EMNLP)}, pages 5811--5826, Online. Association
  for Computational Linguistics.

\bibitem[{Raffel et~al.(2020)Raffel, Shazeer, Roberts, Lee, Narang, Matena,
  Zhou, Li, and Liu}]{raffel2020exploring}
Colin Raffel, Noam Shazeer, Adam Roberts, Katherine Lee, Sharan Narang, Michael
  Matena, Yanqi Zhou, Wei Li, and Peter~J Liu. 2020.
\newblock Exploring the limits of transfer learning with a unified text-to-text
  transformer.
\newblock \emph{The Journal of Machine Learning Research}, 21(1):5485--5551.

\bibitem[{Rahman et~al.(2019)Rahman, Fookes, Baktashmotlagh, and
  Sridharan}]{multicomponent}
Mohammad~Mahfujur Rahman, Clinton Fookes, Mahsa Baktashmotlagh, and Sridha
  Sridharan. 2019.
\newblock Multi-component image translation for deep domain generalization.
\newblock In \emph{2019 IEEE Winter Conference on Applications of Computer
  Vision (WACV)}, pages 579--588.

\bibitem[{Rashkin et~al.(2023)Rashkin, Nikolaev, Lamm, Aroyo, Collins, Das,
  Petrov, Tomar, Turc, and Reitter}]{ais}
Hannah Rashkin, Vitaly Nikolaev, Matthew Lamm, Lora Aroyo, Michael Collins,
  Dipanjan Das, Slav Petrov, Gaurav~Singh Tomar, Iulia Turc, and David Reitter.
  2023.
\newblock {Measuring Attribution in Natural Language Generation Models}.
\newblock \emph{Computational Linguistics}, 49(4):777--840.

\bibitem[{Schuster et~al.(2021)Schuster, Fisch, and
  Barzilay}]{schuster-etal-2021-get}
Tal Schuster, Adam Fisch, and Regina Barzilay. 2021.
\newblock \href {https://doi.org/10.18653/v1/2021.naacl-main.52} {Get your
  vitamin {C}! robust fact verification with contrastive evidence}.
\newblock In \emph{Proceedings of the 2021 Conference of the North American
  Chapter of the Association for Computational Linguistics: Human Language
  Technologies}, pages 624--643, Online. Association for Computational
  Linguistics.

\bibitem[{Thorne et~al.(2018)Thorne, Vlachos, Cocarascu, Christodoulopoulos,
  and Mittal}]{thorne-etal-2018-fact}
James Thorne, Andreas Vlachos, Oana Cocarascu, Christos Christodoulopoulos, and
  Arpit Mittal. 2018.
\newblock \href {https://doi.org/10.18653/v1/W18-5501} {The fact extraction and
  {VER}ification ({FEVER}) shared task}.
\newblock In \emph{Proceedings of the First Workshop on Fact Extraction and
  {VER}ification ({FEVER})}, pages 1--9, Brussels, Belgium. Association for
  Computational Linguistics.

\bibitem[{Tobin et~al.(2017)Tobin, Fong, Ray, Schneider, Zaremba, and
  Abbeel}]{domainrandomization}
Josh Tobin, Rachel Fong, Alex Ray, Jonas Schneider, Wojciech Zaremba, and
  Pieter Abbeel. 2017.
\newblock Domain randomization for transferring deep neural networks from
  simulation to the real world.
\newblock In \emph{2017 IEEE/RSJ International Conference on Intelligent Robots
  and Systems (IROS)}, pages 23--30.

\bibitem[{Vu et~al.(2021)Vu, Luong, Le, Simon, and Iyyer}]{vu-etal-2021-strata}
Tu~Vu, Minh-Thang Luong, Quoc Le, Grady Simon, and Mohit Iyyer. 2021.
\newblock \href {https://doi.org/10.18653/v1/2021.emnlp-main.462} {{ST}ra{TA}:
  Self-training with task augmentation for better few-shot learning}.
\newblock In \emph{Proceedings of the 2021 Conference on Empirical Methods in
  Natural Language Processing}, pages 5715--5731, Online and Punta Cana,
  Dominican Republic. Association for Computational Linguistics.

\bibitem[{Wang et~al.(2020)Wang, Cho, and Lewis}]{wang-etal-2020-asking}
Alex Wang, Kyunghyun Cho, and Mike Lewis. 2020.
\newblock \href {https://doi.org/10.18653/v1/2020.acl-main.450} {Asking and
  answering questions to evaluate the factual consistency of summaries}.
\newblock In \emph{Proceedings of the 58th Annual Meeting of the Association
  for Computational Linguistics}, pages 5008--5020, Online. Association for
  Computational Linguistics.

\bibitem[{Wang et~al.(2021)Wang, Lan, Liu, Ouyang, and
  Qin}]{domaingeneralsurvey}
Jindong Wang, Cuiling Lan, Chang Liu, Yidong Ouyang, and Tao Qin. 2021.
\newblock Generalizing to unseen domains: A survey on domain generalization.
\newblock In \emph{Proceedings of the Thirtieth International Joint Conference
  on Artificial Intelligence, {IJCAI-21}}, pages 4627--4635. International
  Joint Conferences on Artificial Intelligence Organization.
\newblock Survey Track.

\bibitem[{Williams et~al.(2018)Williams, Nangia, and Bowman}]{mnli}
Adina Williams, Nikita Nangia, and Samuel Bowman. 2018.
\newblock A broad-coverage challenge corpus for sentence understanding through
  inference.
\newblock In \emph{Proceedings of the 2018 Conference of the North American
  Chapter of the Association for Computational Linguistics: Human Language
  Technologies, Volume 1 (Long Papers)}, pages 1112--1122.

\bibitem[{Zhang et~al.(2019)Zhang, Baldridge, and He}]{zhang-etal-2019-paws}
Yuan Zhang, Jason Baldridge, and Luheng He. 2019.
\newblock \href {https://doi.org/10.18653/v1/N19-1131} {{PAWS}: Paraphrase
  adversaries from word scrambling}.
\newblock In \emph{Proceedings of the 2019 Conference of the North {A}merican
  Chapter of the Association for Computational Linguistics: Human Language
  Technologies, Volume 1 (Long and Short Papers)}, pages 1298--1308,
  Minneapolis, Minnesota. Association for Computational Linguistics.

\bibitem[{Zhou et~al.(2020)Zhou, Yang, Hospedales, and
  Xiang}]{novelvisiondomains}
Kaiyang Zhou, Yongxin Yang, Timothy Hospedales, and Tao Xiang. 2020.
\newblock Learning to generate novel domains for domain generalization.
\newblock In \emph{Computer Vision -- ECCV 2020}, pages 561--578, Cham.
  Springer International Publishing.

\end{thebibliography}

\clearpage
\pagebreak

\appendix
\appendix

\section{Seed Examples for Prompting}
\label{sec:seed_examples}
We provide the seed examples for prompting FLAN-PaLM2 L (Unicorn) to generate premises in Table \ref{tab:seed_triples}.

\begin{table*}[ht]
\centering
\begin{tabular}{|p{.1\linewidth}|p{.1\linewidth}|p{0.77\linewidth}|}
\hline
\textbf{Domain} & \textbf{Length} & \textbf{Text} \\
\hline
news headlines & short & Congress approves debt deal, averting a US default \\
\hline
news headlines & short & Man airlifted to hospital from Skye beauty spot \\
\hline
news & short & Expectations were set high by the WSC concerning what the event would do for upcoming Indian entrepreneurs. \\
\hline
news & short & But despite high promises, it didn’t take long for the first day of the convention to be plunged into chaos. \\
\hline
shopping reviews & paragraph & Good value for the seventy eight dollars that I paid for it. easy to change the filter. Quite on high. Haven't had it long enough to say how well it filters the air but I can see lint and dust on the filter pre screen. And I've only had it nine days I think. Love that I can turn the lights off. \\
\hline
shopping reviews & short & my first impressions are that's the Google Pixel 7 is a nice phone, BUT not as good as the moto g power in terms of ease of use and functionality. \\
\hline
shopping reviews & short & Battery has yet to be determined on the Pixel, but from a full charge, I'm down to 56\% after 2 hours of use. \\
\hline
wikipedia & paragraph & Alfred was baptised by Frederick Cornwallis, Archbishop of Canterbury, in the Great Council Chamber at St James's Palace on 21 October 1780. His godparents were his elder siblings George, Prince of Wales; Prince Frederick; and Charlotte, Princess Royal. Alfred was a delicate child. \\
\hline
wikipedia & short & The premise of Two Hundred Rabbits was based on a dream that author Lonzo Anderson had after reading a French folk tale. \\
\hline
movie reviews & paragraph & As usual, James Cameron shows us his creative genius. The story is very different from the first, and I don't want to give out any story until you've seen it. It is worth watching, and if you own the first it is also worth buying. My only complaint, and it is BIG, is it turns out to only be in 480p resolution...not even 1080p or 4K. It looks good if you play it in YouTube, but still. It should be in 4K. \\
\hline
movie reviews & short & The actor portraying Mr. Darcy had no concept of the kind of man Darcy is or his nature. \\
\hline
place reviews & paragraph & Beautiful space which is nicely a bit secluded from the hussle at coal drop but still easy to reach. Wines were excellent, cheeses delicious, food great, and cocktails outstanding. Folks were kind and professional. Crowd was elegant but relaxed. \newline \newline Amazed they just opened three days ago, they operate like they have been at it forever. Loved every minute! \\
\hline
place reviews & short & The steep stairs need to be negotiated with caution especially after indulging in bout of revelry. \\
\hline
place reviews & short & I waited an hour. The doctor was terribly stressed. She didn't answer questions. \\
\hline
twitter & short & Sevilla is {\bf Red and White} \heart \\
\hline
twitter & short & Lil X just asked if there are police cats, since there are police dogs :)) \\
\hline
reddit post & paragraph & Hey there everyone! I often see people asking where to start when getting into prog metal, so I thought instead of answering every one of them individually I'd make a list. I'm not going into too much depth because otherwise this will become endless, but I'll try to give a brief explanation of all styles I'm going over. So let's get started! \\
\hline
reddit post & short & I am someone who hates doing laundry. \\
\hline
\end{tabular}
\caption{Seed Examples for Prompting.}
\label{tab:seed_triples}
\end{table*}

\section{Running time and Hyper-parameter Details of Prompt Tuning Experiments}
\label{sec:hyper-parameters-palm}

For prompt-tuning of FLAN-PaLM 540B, we used an input length and output length of $512$. We tuned $100$ prompt embeddings and used them during inference. We used a learning rate of $0.3$ and did not use dropout. We trained with a batch size of $16$ for $24,544$ steps that is equivalent to one epoch on the MNLI dataset with $392,702$ training examples. The prompt-tuning took around 110 hours to complete with 256 Cloud TPU v4 chips.

\section{Hyper-parameter Details of T5 Fine-tuning Experiments}
\label{sec:hyper-parameters-t5}

For training T5 models (both binary and 3-way), we tuned learning rates $\in \{5e-4,1e-4,5e-5\}$ and fine-tuned with batch size of $32$ for $50$K steps. We checkpoint every $1K$ steps for early stopping. We use a dropout rate of $0.1$. We trained with an input length of $512$. During inference for factual consistency evaluation (Section \ref{sec:true}) and NLI benchmarks (Section \ref{sec:3-way}), we used an input length of $1024$ and $512$ respectively.

We report the best selected hyper-parameters for T5 binary and 3-way models in Table \ref{tab:t5_binary_hparams} and Table \ref{tab:t5_hparams}.

\begin{table*}[htbp]
    \centering
    \small
    \begin{tabular}{|l|c|c|c|}
        \hline
        \textbf{Dataset/Model} & \textbf{T5 small} & \textbf{T5 large} & \textbf{T5 XXL} \\
        \hline
        \textbf{MNLI} & lr=5e-4, steps=40K & lr=5e-4, steps=10K & lr=5e-4, steps=15K \\
        \hline
        \textbf{ANLI} & lr=5e-4, steps=5K & lr=5e-4, steps=20K & lr=5e-4, steps=40K \\
        \hline
        \textbf{WANLI} & lr=5e-4, steps=15K & lr=5e-4, steps=10K & lr=1e-4, steps=50K \\
        \hline
        \textbf{GNLI} & lr=5e-4, steps=40K & lr=5e-4, steps=20K & lr=5e-5, steps=40K \\
        \hline
        \textbf{MNLI + GNLI} & lr=5e-4, steps=20K & lr=5e-4, steps=35K & lr=5e-5, steps=25K \\
        \hline
        \textbf{ANLI + GNLI} & lr=5e-4, steps=50K & lr=5e-4, steps=20K & lr=5e-4, steps=50K \\
        \hline
        \textbf{WANLI + GNLI} & lr=5e-4, steps=45K & lr=5e-4, steps=30K & lr=5e-5, steps=50K \\
        \hline
    \end{tabular}
    \caption{Best selected hyper-parameters for T5 binary models. We report learning rates (lr) and the number of steps.}
    \label{tab:t5_binary_hparams}
\end{table*}

\begin{table*}[ht!]
\centering
\small
\begin{tabular}{|c|c|c|c|}
\hline
& \textbf{T5 small} & \textbf{T5 large} & \textbf{T5 XXL} \\
\hline
\textbf{MNLI} & lr=5e-4, steps=40K & lr=5e-4, steps=10K & lr=5e-5, steps=35K \\ \hline
\textbf{ANLI} & lr=5e-4, steps=45K & lr=5e-4, steps=10K & lr=5e-4, steps=15K \\ \hline
\textbf{WANLI} & lr=5e-4, steps=10K & lr=5e-4, steps=10K & lr=5e-5, steps=15K \\ \hline
\textbf{GNLI} & lr=5e-4, steps=50K & lr=5e-4, steps=15K & lr=5e-5, steps=35K \\ \hline
\textbf{MNLI + GNLI} & lr=5e-4, steps=25K & lr=5e-4, steps=45K & lr=5e-4, steps=45K \\ \hline
\textbf{ANLI + GNLI} & lr=5e-4, steps=05K & lr=5e-4, steps=45K & lr=5e-5, steps=45K \\ \hline
\textbf{WANLI + GNLI} & lr=5e-4, steps=50K & lr=5e-4, steps=35K & lr=5e-5, steps=35K \\
\hline
\end{tabular}
\caption{Best selected hyper-parameters for T5 3-way classification models. We report learning rates (lr) and the number of steps.}
\label{tab:t5_hparams}
\end{table*}

\end{document}